\documentclass[12pt]{article}

\usepackage{scicite}
\usepackage{times}
\usepackage{graphics} 
\usepackage{epsfig} 
\usepackage{mathptmx} 
\usepackage{times} 
\usepackage{amsmath} 
\usepackage{amssymb}  

\usepackage{tabularx}
\usepackage{booktabs}
\usepackage{multirow}

\usepackage{siunitx}
\usepackage{xcolor}
\usepackage[normalem]{ulem}
\usepackage{comment}

\newcommand{\refig}[1] {Fig.~\ref{#1}}
\usepackage{gensymb}

\newenvironment{sciabstract}{%
\begin{quote} \bf}
{\end{quote}}

\title{
Terradynamics and design of tip-extending robotic anchors
}

\author
{Deniz Kerimoglu$^{1\ast}$, Nicholas D. Naclerio$^{2\ast}$, Sean Chu$^{3}$, Andrew Krohn$^{3}$,\\ Vineet Kupunaram$^{3}$, Alexander Schepelmann$^{4}$, Daniel I. Goldman$^{1}$ and\\ Elliot W. Hawkes$^{2\ast\ast}$ \\
\\
\normalsize{$^{1}$School of Physics, Georgia Institute of Technology, Atlanta, GA 30332, USA}\\
\normalsize{$^{2}$ Department of Mechanical Engineering, University of California, Santa Barbara,}\\
\normalsize{Santa Barbara, CA 93106, USA}\\
\normalsize{$^{3}$ College of Engineering, University of California, Santa Barbara,}\\
\normalsize{Santa Barbara, CA 93106, USA}\\
\normalsize{$^{4}$ NASA Glenn Research Center, Cleveland, OH 44135, USA}\\
\normalsize{$^{\ast}$ These authors contributed equally.}\\
\normalsize{$^{\ast\ast}$To whom correspondence should be addressed; E-mail:  ewhawkes@ucsb.edu.}
}

\date{}

\begin{document} 
\baselineskip24pt
\maketitle 

\begin{sciabstract}

Most engineered pilings require substantially more force to be driven into the ground than they can resist during extraction. This requires relatively heavy equipment for insertion, which is problematic for anchoring in hard-to-access sites, including in extraterrestrial locations. In contrast, for tree roots, the external reaction force required to extract is much greater than required to insert--little more than the weight of the seed initiates insertion. This is partly due to the mechanism by which roots insert into the ground: tip extension. 
Proof-of-concept robotic prototypes have shown the benefits of using this mechanism, but a rigorous understanding of the underlying granular mechanics and how they inform the design of a robotic anchor is lacking.
Here, we study the terradynamics of tip-extending anchors compared to traditional piling-like intruders, develop a set of design insights, and apply these to create a deployable robotic anchor. Specifically, we identify that to increase an anchor's ratio of extraction force to insertion force, it should: (i) extend beyond a critical depth; (ii) include hair-like protrusions; (iii) extend near-vertically, and (iv) incorporate multiple smaller anchors rather than a single large anchor.
Synthesizing these insights, we developed a lightweight, soft robotic, root-inspired anchoring device that inserts into the ground with a reaction force less than its weight. We demonstrate that the 300 g device can deploy a series of temperature sensors 45 cm deep into loose Martian regolith simulant while anchoring with an average of 120 N, resulting in an anchoring-to-weight ratio of 40:1.\\

\end{sciabstract}

\section*{INTRODUCTION}
\label{sec:Intro}
Both redwood trees and cellphone towers require strong roots or foundations to anchor against powerful environmental forces such as storms and earthquakes, but their anchors are deployed in fundamentally different ways. Engineered anchors are typically pushed into the ground by pile driving, excavation, or drilling, processes that require heavy equipment \cite{hwang2001ground,leung2000behavior,shuter1989application}. In contrast, tree roots require almost no external reaction force to begin insertion--little more than the weight of their seed suffices \cite{bengough2016root}--
yet, the resulting structure can anchor against tremendous uprooting forces \cite{edmaier2014influence,ennos1990anchorage,ghidey1997plant}. A key feature that distinguishes biological roots from engineered anchors is tip extension, a mechanism in which only the tip advances into the medium \cite{ennos2000mechanics,kolb2017physical}, and may be responsible for this minimal insertion force. 

Tip-extending anchors, capable of inserting with small reaction forces like biological roots, could have important applications in environments where insertion forces are difficult to produce, such as in reduced gravity and remote locations where logistical costs of transporting massive, energy-intensive equipment are prohibitively high \cite{bar2009drilling}. Example applications include drilling for volatiles on the moon \cite{kleinhenz2015impact}, deploying temperature sensors on Mars \cite{lipatov2023temperature}, and mining asteroids \cite{zacny2013asteroids}. As a result, there is a pressing need for anchoring strategies that minimize insertion reaction force while maximizing extraction strength.

\begin{figure}
    \centering
    \includegraphics[width=.5\columnwidth]{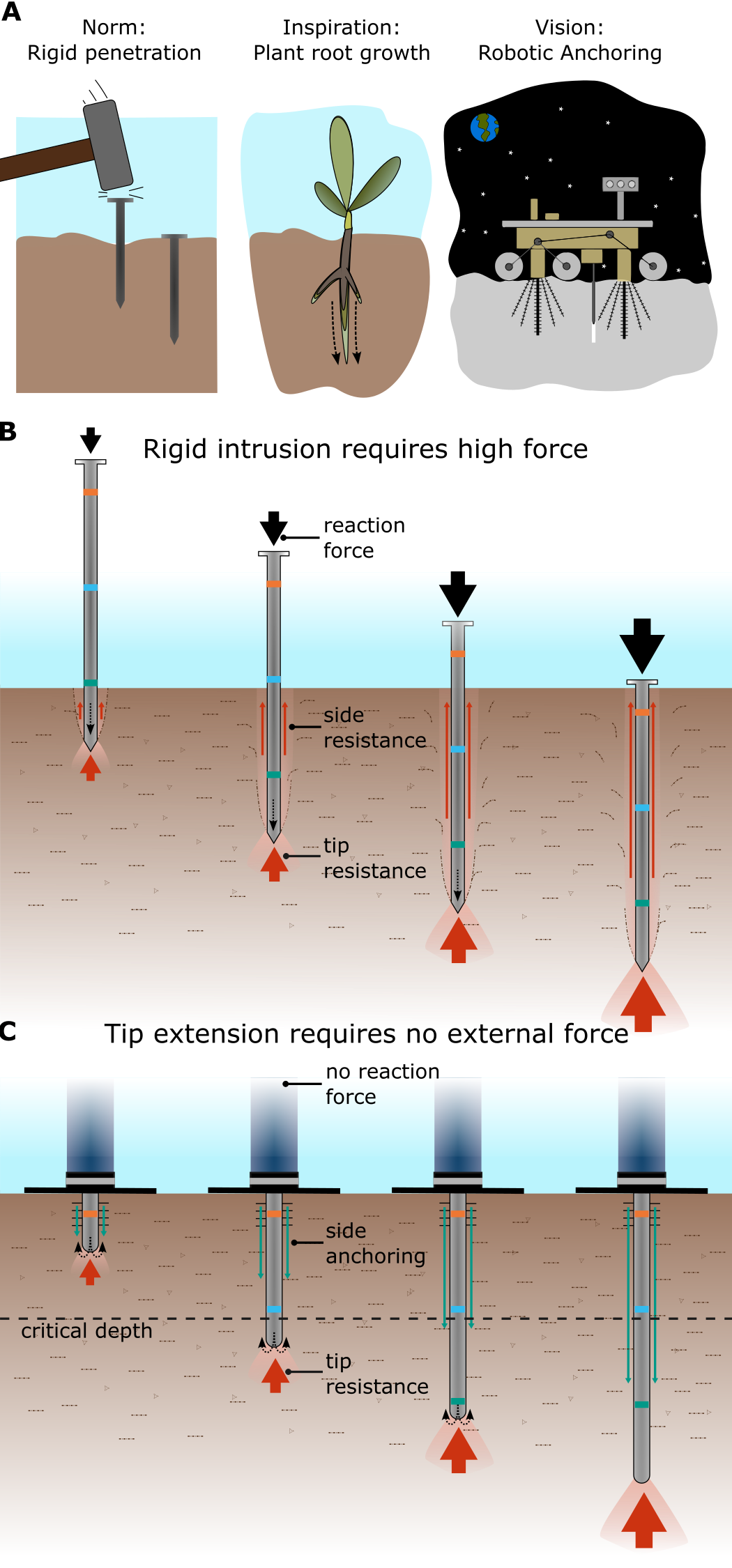}
    \caption{ \textbf{Overview of the motivation and mechanisms of soft robotic anchor.} (\textbf{A}) Rigid insertion and (\textbf{B}) plant root growth are fundamentally different mechanisms of anchoring in the ground. (\textbf{C}) Growth offers advantages in remote and low-gravity environments for robotic anchoring. (\textbf{D}) Rigid insertion requires high force because it must overcome both tip and side resistances. (\textbf{E}) Conversely, tip-extension enables ``zero-reaction-force self-anchoring'' because anchoring forces along the sides of the device counter tip resistance beyond a critical depth.}
    \label{fig:int_grow}
\end{figure}

Prior work has explored topics related to robotic tip-extending anchors.
Work in biology has explored tip-extending root growth and anchoring through experiments and simulations (e.g., \cite{koren2024analysis,pirrone2023methodology,stubbs2019general,yang2017root,yang2018analyzing}).
Prior engineering work has demonstrated burrowing into granular media with tip extension \cite{naclerio2018soft,sadeghi2017toward,sadeghi2014novel,sachin2025beerootbot}, measured the pullout capacity of buried root-inspired anchors \cite{dyson2014pull}, and developed mechanisms and simulated probes that increase anchoring force \cite{martinez2020evaluation,chen2021modeling,dupuy2005numerical}. 
In previous work from the authors \cite{naclerio2021controlling}, a soft, tip-extending burrowing robot showed anchoring behavior during horizontal growth, but anchoring performance was neither explored nor quantified. Sadeghi et al. \cite{sadeghi2013robotic} reported a reduction in insertion force using a skin-everting tube, but did not investigate the anchoring response in detail. 
As such, a rigorous exploration of the terradynamic principles underlying robotic tip extension and their application to the design of a robotic anchor is lacking. 

Here, we first studied the terradynamics of tip-extending anchorage in dry granular media, comparing the behaviors of a soft robotic, tip-extending device with those of a rigid intruder.
We found that tip extension reduces insertion forces and increases extraction forces, and we examined how diameter and angle affected these forces.
We also discovered that when the tip extender was not constrained vertically during insertion, ``zero-reaction-force self-anchoring" occurs, in which the reaction force during insertion drops to zero at a critical depth.
Second, based on the results of this study, we developed a set of design insights for robotic anchors. 
Specifically, we found that for a tip-extending anchor to increase the ratio between its extraction and insertion force, it should: (i) extend beyond a critical depth, (ii) include hair-like protrusions, (iii) extend near-vertically, and (iv) incorporate multiple smaller anchors rather than a single large anchor. 
Third, guided by these insights, we designed and built a lightweight, soft-robotic, root-inspired, self-anchoring device that burrows into the ground without external insertion force. The 300 g device deployed temperature sensors 45 cm deep into Martian regolith simulant and anchored with an average of 120 N, an anchoring-to-weight ratio of 40:1.

\section*{RESULTS}

\subsection*{Tip-extension Terradynamics Experiments}
\label{sec:terradynamics}

We conducted a series of experiments to investigate the terradynamics of tip extension. First, we measured the force and movement of particles during insertion and extraction. Second, we conducted a set of anchoring tests in which we varied the angle of insertion and the diameter of the anchor. Third, we studied the phenomenon of ``zero-reaction-force self-anchoring,'' in which the insertion force drops to zero beyond a critical depth (Fig. \ref{fig:int_grow}).

\paragraph*{Tip extension reduces insertion force and increases extraction force}
\label{sec:position_controled_forces}

\begin{figure}[!ht]
    \centering
    \includegraphics[width=1\columnwidth]{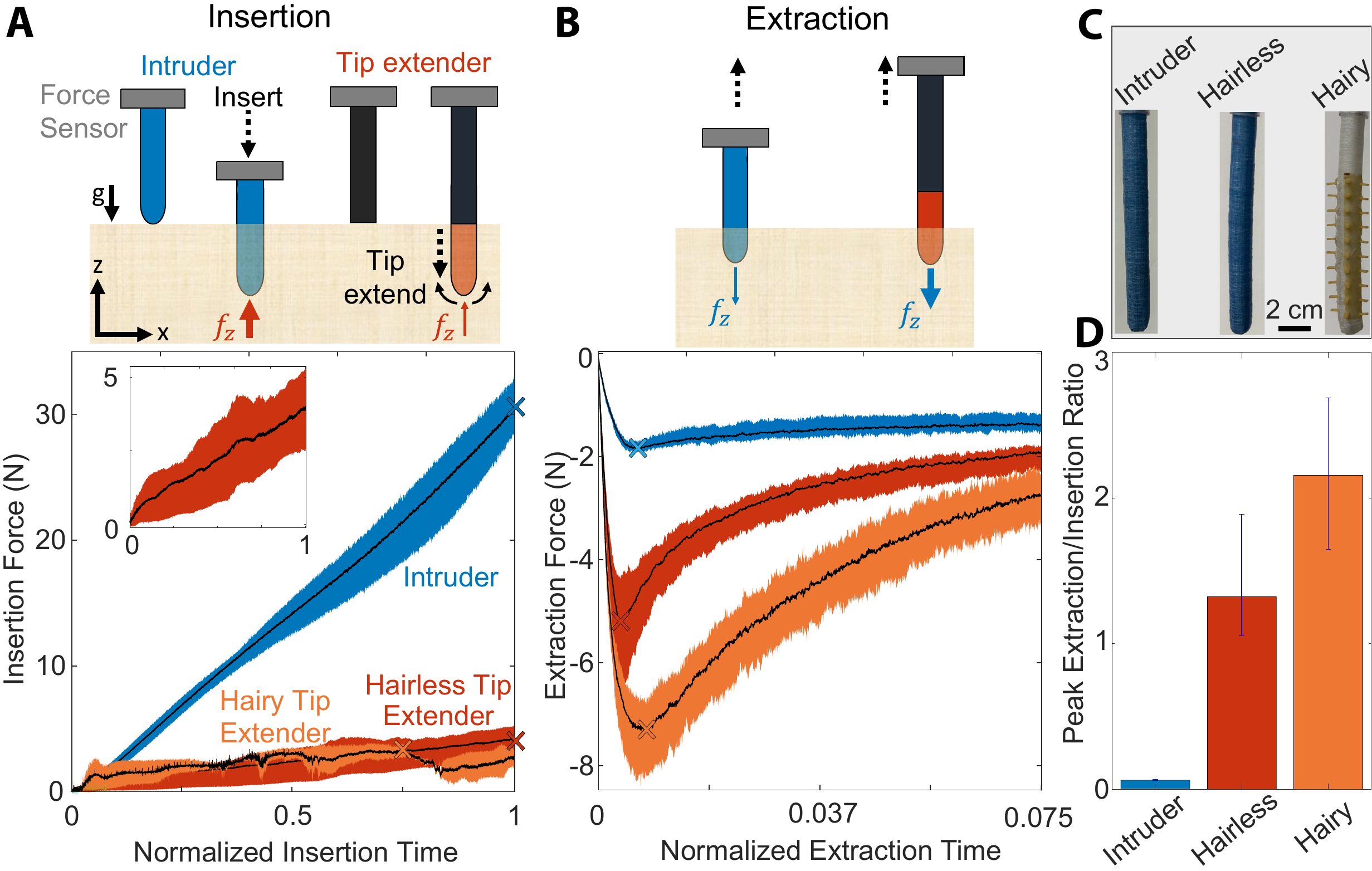}
    \caption{\textbf{Insertion and extraction forces measured at the base of rigid intruder (blue), hairless (red), and hairy tip extender (orange) in sand.} 
    (\textbf{A}) 
    Insertion force as a function of normalized insertion time to a depth of 15 cm, where time = 0 represents no insertion and time = 1 indicates full insertion. Depth during insertion was not directly measured. The average times for full insertion were 13.5 s for the rigid intruder, 10.3 s for the hairless, and 12.5 s for the hairy tip extender. The mean standard deviations of the shaded regions are 1.1 N, 0.85 N, and 1.08 N, respectively. \textit{Inset:} Zoom of hairless tip extender.
    (\textbf{B}) 
    Extraction force as a function of normalized time. The average extraction time for all the devices was 13.5 s. The mean standard deviations of the shaded regions are 0.05 N, 0.09 N, and 0.17 N, respectively. 
    (\textbf{C}) The devices used for insertion and extraction experiments: rigid intruder, hairless tip extender, and hairy tip extender. (\textbf{D}) Ratio of maximum extraction to maximum insertion force. The location of maximum forces is marked by an ``x'' in (\textbf{A}) and (\textbf{B}). The error bars represent the lower and upper bounds of the peak extraction-to-insertion force ratios.}
    \label{fig:int_pull_sand}
\end{figure}

We measured forces and particle velocities of a rod-shaped rigid intruder, a hairless tip extender, and a hairy tip extender during insertion and extraction in loose sand. The insertion force for the rigid intruder increased nearly linearly with depth, as expected. The tip extenders followed a similar trend, but with approximately 10 times lower slope, resulting in the maximum insertion force on the intruder being approximately 10 times greater than that on either of the tip extenders (\refig{fig:int_pull_sand}A). 
Conversely, the maximum extraction force of the intruder was 2.5 and 3.5 times lower than the hairless and hairy tip extenders, respectively (\refig{fig:int_pull_sand}B). 
We observed that the intruder exhibited a relatively flat force profile after failure, characteristic of a dynamic-friction-dominated behavior (blue curve), whereas the tip extenders exhibited a substantial decrease in force after reaching a peak, more closely resembling ductile yielding behavior (red and orange curves).
This drop-off is more rapid in the hairless device. 
For the ratio of extraction force to insertion force, both tip extenders had higher peak extraction forces than peak insertion forces, whereas the intruder had a higher peak insertion force than extraction force (Fig. \ref{fig:int_pull_sand}D). 

\begin{figure}[!h]
    \centering
\includegraphics[width=.69\linewidth]{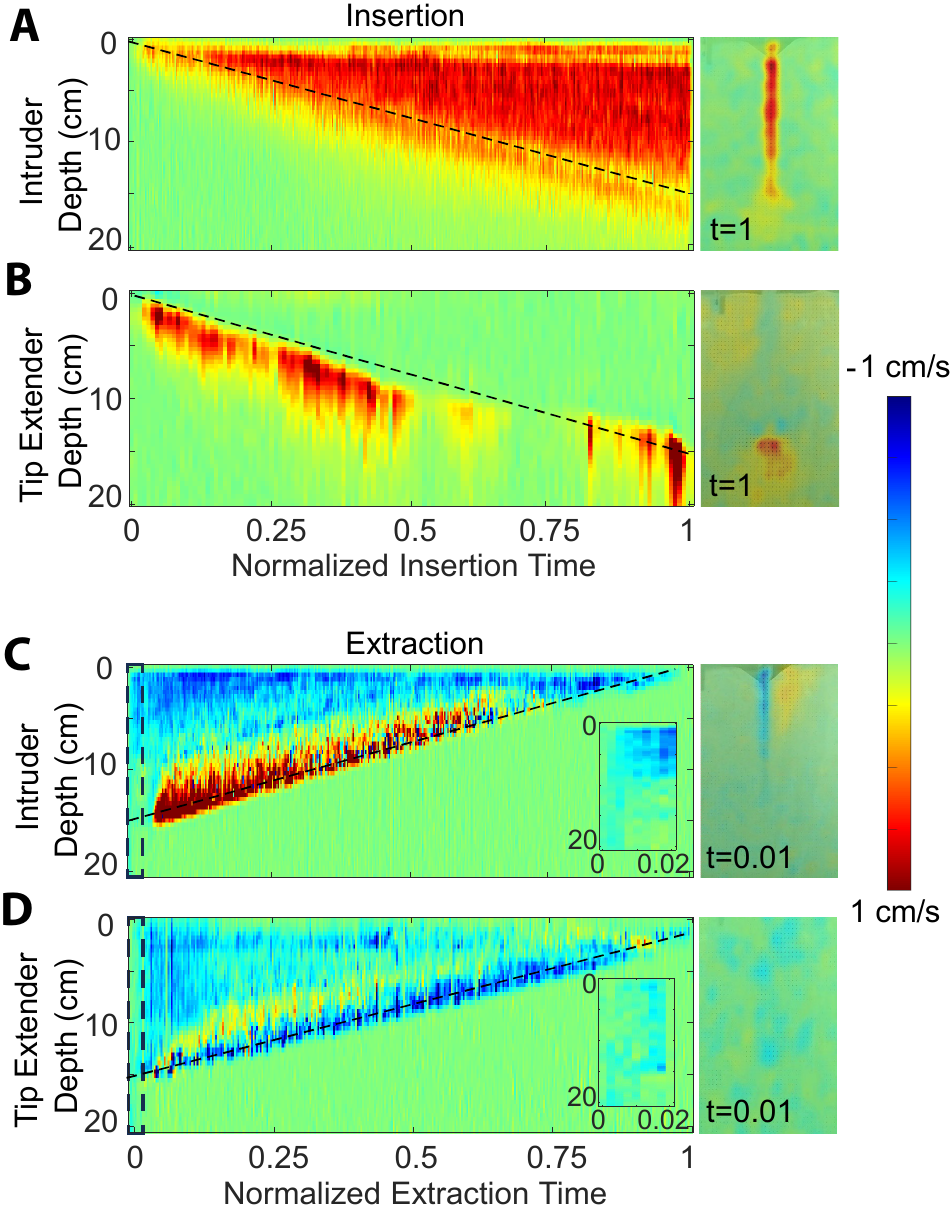}
    \caption{\textbf{Particle Image Velocimetry (PIV) reveals granular flow dynamics during insertion and extraction of an intruder (top row), hairless tip extender (bottom row).} Heatmaps are constructed by sequentially concatenating vertical slices captured at the midpoint over the entire duration of insertion and extraction. The experiments were conducted in a confined sand testbed. The color maps represent the magnitude and direction of particle velocities, and the dashed diagonal lines represent the approximate tip position of both devices. The dashed rectangles mark the onset of extraction, which is shown in detail in the enlarged subset figures at the bottom right. (\textbf{A}) During intruder insertion, the nearby particles are continuously displaced. (\textbf{B}) In contrast, during tip extension, the particles around the tip are displaced, while the sand above the growing region remains nearly stationary. (\textbf{C-D}) At the onset of extraction (inset), the tip extender (\textbf{D}) exhibits less particle movement than the rigid intruder (\textbf{C}).}
    \label{fig:PIV}
\end{figure}

We tracked the movement of sand grains using particle image velocimetry (PIV) during the insertion and extraction of devices to gain further insight into the terradynamics of tip-extension (Fig. \ref{fig:PIV}). Sand particles were confined between two clear plates, spaced to match the device diameters. This setup constrained the devices' motion to a quasi-2D plane and enabled clear observation of the particles near the side of the devices touching the transparent wall.
During insertion of the rigid intruder, particles near the tip and sides were continuously displaced and pushed downward as the intruder penetrated deeper (Fig. \ref{fig:PIV}A). In contrast, for the tip extender, only the particles near the tip had substantial velocity, while the sand above the growing region remained nearly stationary (Fig. \ref{fig:PIV}B). The immobile region of sand between the surface and the tip lengthened as the device grew deeper. Our rigid insertion results are consistent with soil mechanics literature, showing that resistive forces on an intruding body arise from continuous material displacement, shearing, and loading around it\cite{terzaghi1943theoretical,salgado2012mechanics,pirrone2022historical}. In contrast, tip extension localizes material displacement to the tip region, reducing the volume of particles that must be mobilized to accommodate the intruding body and leading to lower insertion resistance.

The extraction PIV revealed particle rearrangement differences between the two devices, despite their identical extraction geometry (Fig. \ref{fig:PIV}\textbf{C–D}). Specifically, the extraction of the rigid intruder induced substantial material yielding and a downward flow of particles (red regions), in contrast to the tip extender, which exhibited a more uniform and upward particle displacement. To study the extraction mechanics, we focused on the onset of extraction of both devices (subsets in  Fig. \ref{fig:PIV} \textbf{C-D}), where peak extraction force occurs. During this phase, the particles surrounding the tip extender showed less mobility compared to those around the rigid intruder. This decreased particle mobility may result in greater resistance to extraction. The difference in particle mobility potentially originates from microscopic particle interactions established by devices during their insertion. See the discussion for details.

\begin{figure} [!h]
    \centering
    \includegraphics[width=.78\columnwidth]{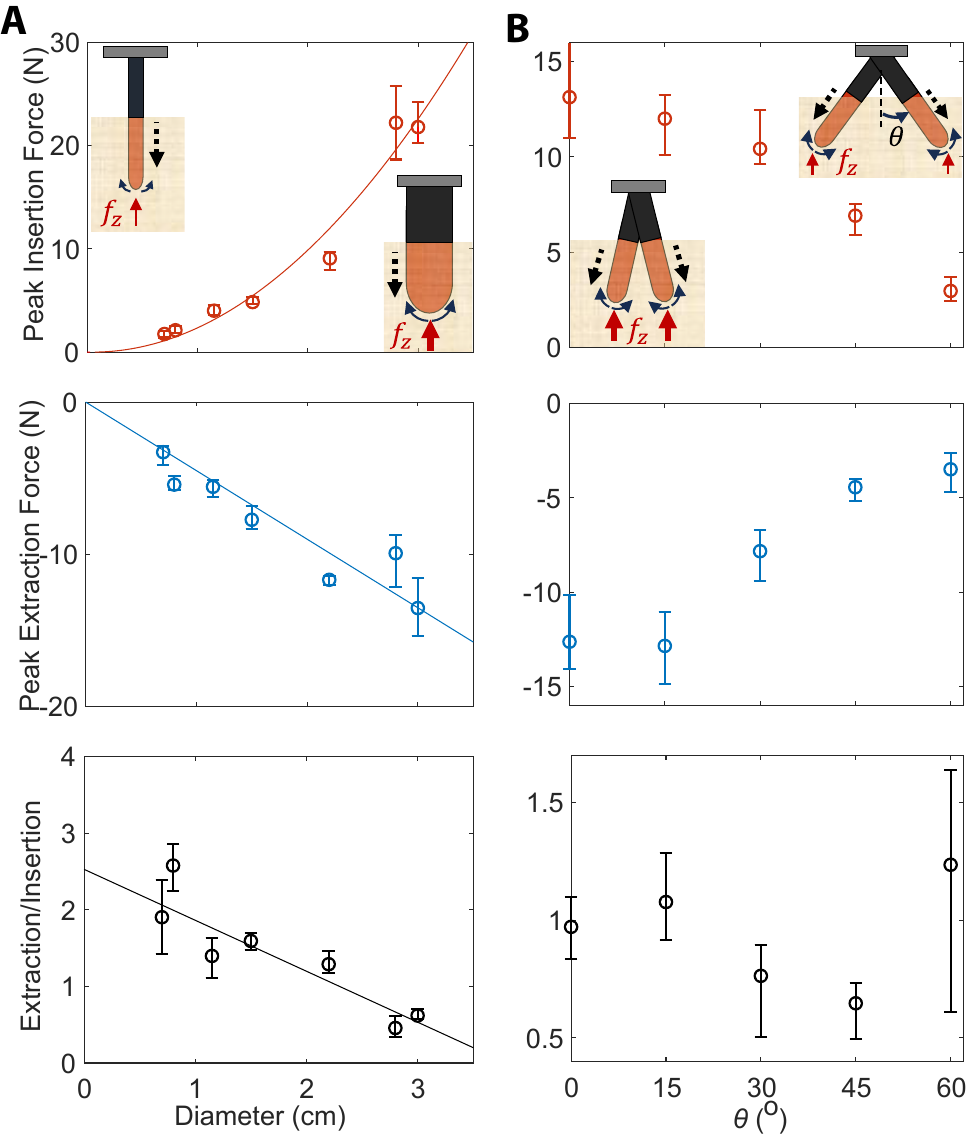}
    \caption{\textbf{The effects of diameter and angle on insertion and extraction}. (\textbf{A}) Resistive forces measured for tip extenders with varying diameter during insertion (top) and extraction (middle), and the corresponding extraction-to-insertion ratio (bottom). Solid curves represent the quadratic and linear fits to the insertion and extraction forces, respectively, and the resulting linear fit to the ratio. (\textbf{B}) Resistive forces measured for a pair of tip extenders with varying insertion angles during insertion (top), extraction forces (middle), and the corresponding extraction-to-insertion ratio (bottom).    }
    \label{fig:dia_scale}
\end{figure}

\paragraph*{Measuring the effects of diameter and angle on insertion and extraction}
\label{sec:media}
To study the effect of varying diameter on insertion and extraction forces, we measured the forces of tip extenders with different diameters as they extended to a fixed depth in sand.
Our results in Fig. \ref{fig:dia_scale}A suggest that the extraction forces increase approximately linearly with diameter, while the insertion forces scale approximately quadratically.
Therefore, the extraction-to-insertion force ratio decreases with increasing diameter. We fit solid lines to the experimental data to illustrate the quadratic and linear trends for insertion and extraction forces, respectively.

To investigate the effect of angle, we measured the vertical resistive force experienced by a pair of hairless tip extenders at varying angles with respect to the vertical. As shown in Fig. \ref{fig:dia_scale}B, both the insertion and extraction forces decrease at angles away from vertical, but because the extraction force decreases more rapidly, the extraction-to-insertion force ratio appears to decrease at the angles of 30 and 45 degrees. At the highest angle tested, 60 degrees, the ratio may increase again, but the data has large uncertainty.

\paragraph*{Self-anchoring occurs when the anchor is unconstrained}
\label{sec:force_controlled_experiment}
In all of the above experiments, the tip extender was rigidly mounted below a force gauge, and the insertion force increased with increasing depth. 
However, we discovered that when we removed this constraint, the insertion force dropped to zero beyond a critical depth.
We termed this emergent phenomenon ``zero-reaction-force self-anchoring.'' 
To systematically investigate this, we created an unconstrained, force-controlled measurement by placing a series of weights on the top of the tip extenders as they grew into the sand. To measure insertion force, we recorded the minimum weight required to prevent the tip extender from backing out of the sand, starting at various depths. After growing to a specified depth, we pulled the anchor out with a force gauge to measure the extraction force.

\begin{figure}[!h]
    \centering
    \includegraphics[width=.57\linewidth]{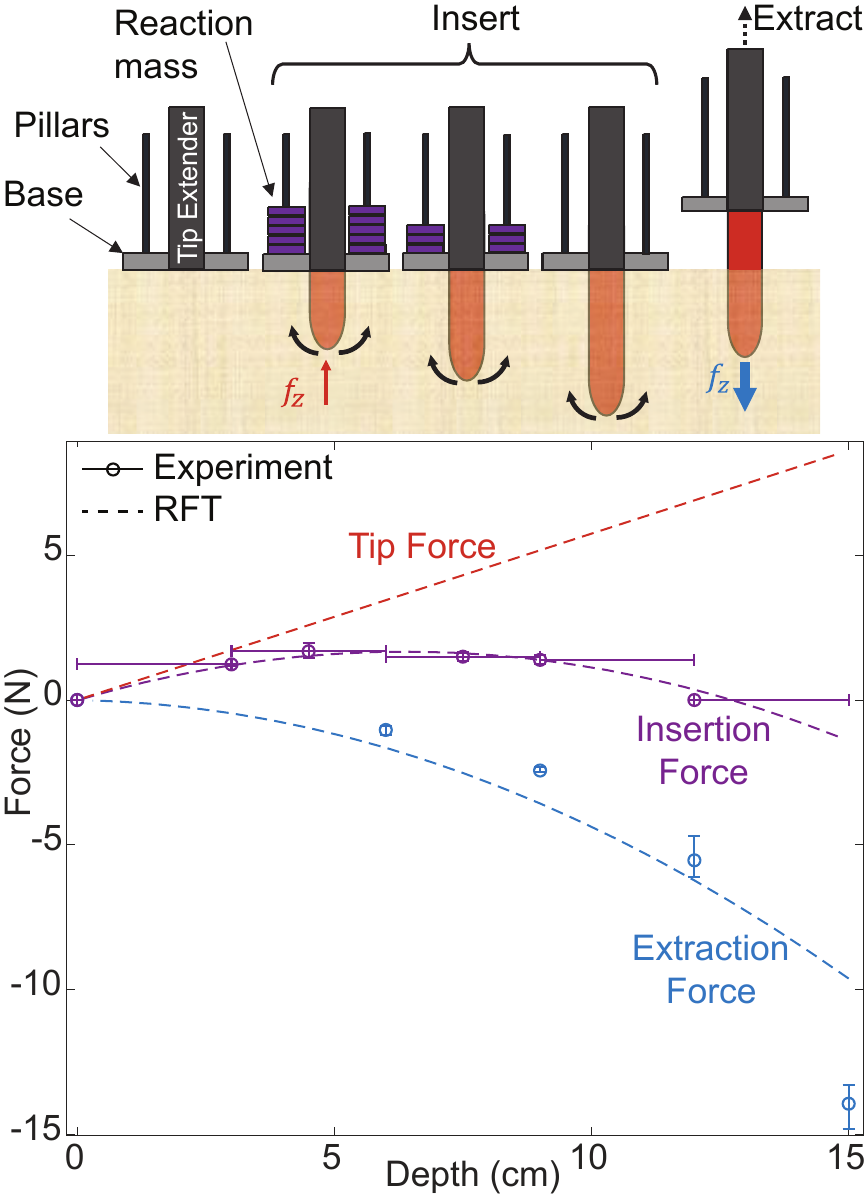}
    \caption{ \textbf{Self-anchoring experiments}.
    Insertion and extraction force of a tip extender in a force-controlled setup in loose sand demonstrates the ``zero-reaction-force self-anchoring" phenomenon.  
    The insertion force is measured as the minimum weight on a tip extender to prevent it from backing out of the sand as it grows deeper. The maximum insertion force was recorded at 3 cm intervals.
    The RFT model is shown as a dashed line.
    Tip force is not directly measured and equals net force minus extraction force. 
    }
    \label{fig:self-anchoring}
\end{figure}

We plot the results in Figure \ref{fig:self-anchoring} and compare to a simple resistive force theory (RFT) model \cite{li2013terradynamics}; both show that the net force approaches zero at a critical length of 12 cm. 
The model predicts this because, according to RFT, there is both resistance to insertion at the tip that scales linearly with depth as well as resistance to extraction along the sides (static with respect to the surrounding substrate) that scales quadratically with depth. 
Thus, RFT predicts that the interplay between these opposing forces should result in a crossover depth beyond which the side anchoring force is greater than the tip resistance, and no net insertion force is needed to grow deeper. See materials and methods for details.

\subsection*{Design insights}
\label{sec:design_insights}

From the results of our terradynamics study, we distilled four key design insights that help increase the extraction-to-insertion ratio of tip-extending anchors.

\paragraph{Insight 1: Anchors should extend beyond a critical depth}
As shown in Fig. \ref{fig:self-anchoring}, the insertion force drops to zero at a critical depth, and if tip extension were to continue, this force would remain zero.
At the same time, the anchoring or extraction force continues to grow quadratically with depth. 
Therefore, substantial increases in extraction-to-insertion ratio can be generated by extending past the critical length of the device. 

\paragraph{Insight 2: Hairs improve anchoring}
As shown in Fig. \ref{fig:int_pull_sand}, hairs increase both the maximum force and work required to extract an anchor without substantially changing the insertion force.
\paragraph{Insight 3: Anchors should extend near-vertically}
\label{sec:angles}

Tip-extending anchors should remain within approximately 15 degrees of vertical. 
Our results in Fig. \ref{fig:dia_scale}B suggest a decrease in extraction force beyond 15 degrees of vertical and possibly a decrease in extraction-to-insertion ratio at 30 and 45 degrees. 

\paragraph{Insight 4: Anchors should have multiple small roots and deploy smaller roots before larger ones}
\label{sec:diameters}
For a given depth, several small anchors have better anchoring performance than one large device. 
This is because, according to the results in Fig. \ref{fig:dia_scale}A, side forces scale with surface area (which scales linearly with diameter, $\propto d$) while tip-resistance scales with cross-sectional area (which scales quadratically with diameter, $\propto d^2$).
Therefore the ratio of side to tip forces increases as the diameter of the anchor decreases ($\propto 1/d$).
As such, for a given cross-section of anchor, it is better to divide the anchor into many small roots, up to the practical limit.
Additionally, heterogeneous root diameters can be advantageous if the smaller diameter roots are deployed first and offer enough anchoring force to deploy larger roots (which offer more extraction force but require more insertion force). 
This sequence should be helpful when seeking to maximize the extraction-to-insertion force ratio; we implemented such a design in the anchor described below.

\subsection*{Soft Robotic Anchoring Device Design }
\label{sec:robot_design}

Leveraging our results and insights from above, we designed and built a lightweight soft robotic, self-anchoring device for subsurface sensor placement and anchoring.
The device features anchors that insert at least 3 times deeper than the critical depth (Insight 1), include hairs (Insight 2), are deployed nearly vertically (Insight 3), and are separated into four roots (three smaller and one larger) that deploy in two stages (Insight 4).

\begin{figure}
    \centering
    \includegraphics[width=0.7\linewidth]{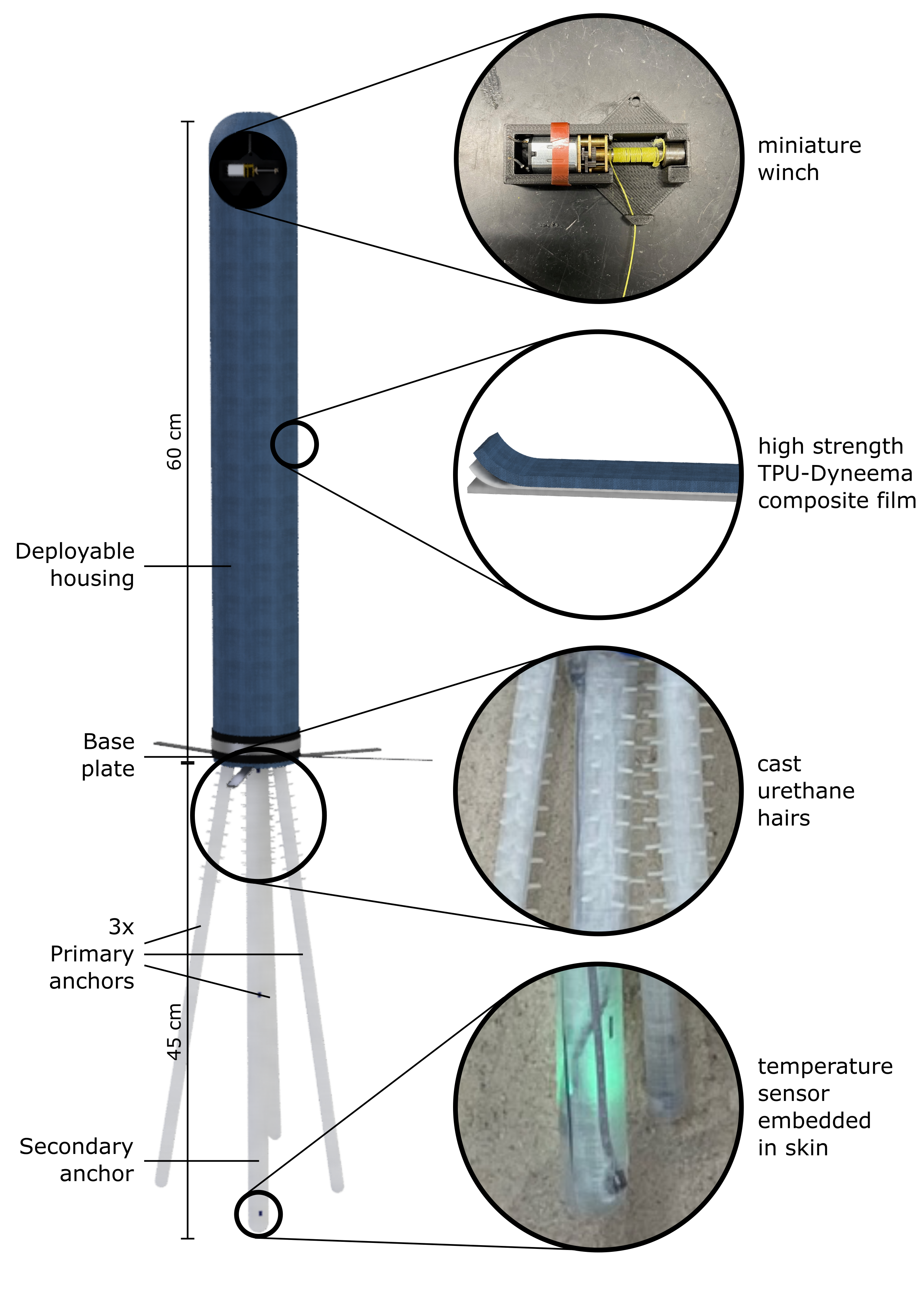}
    \caption{\textbf{The soft robotic anchoring device.} The robot achieves a relatively high anchoring force with a relatively low mass. Its four main components are labeled at left: deployable housing, base plate, primary anchors, and secondary anchor.
    A miniature sequencing winch inside the body controls when the secondary anchor deploys.
    The anchors are made of a lightweight, high strength thermoplastic polyurethane (TPU) and Dyneema composite film. Urethane hairs and temperature sensors are bonded to the skin of the anchors.}
    \label{fig:robot_design}
\end{figure}

The device comprises four main components, all designed to maximize anchoring force while minimizing mass (Figure \ref{fig:robot_design}).
First is a 60 cm tall, deployable, inflatable housing that stores the retracted anchors and motorized sequencing winch that controls the anchor deployment.
Second is a 3D-printed nylon base plate with carbon fiber ``legs" for stability that connects the inflatable housing to four anchors and provides inlets for pressure and wires for the sequencing winch, temperature sensors, and LED for visualization.
Third is a set of three 13 mm diameter, 45 cm long, primary anchors for providing initial anchoring. These small-diameter anchors have a lower insertion force than the larger anchor, but self-anchor and produce enough anchoring force that the large anchor requires no additional external reaction force to anchor.
Fourth is the single 20 mm diameter, 45 cm long, secondary anchor with four 2-wire temperature sensors embedded along its length for measuring subterranean temperature and a green LED attached to its tip for visualization.

\begin{figure}
    \centering
    \includegraphics[width=0.9\linewidth]{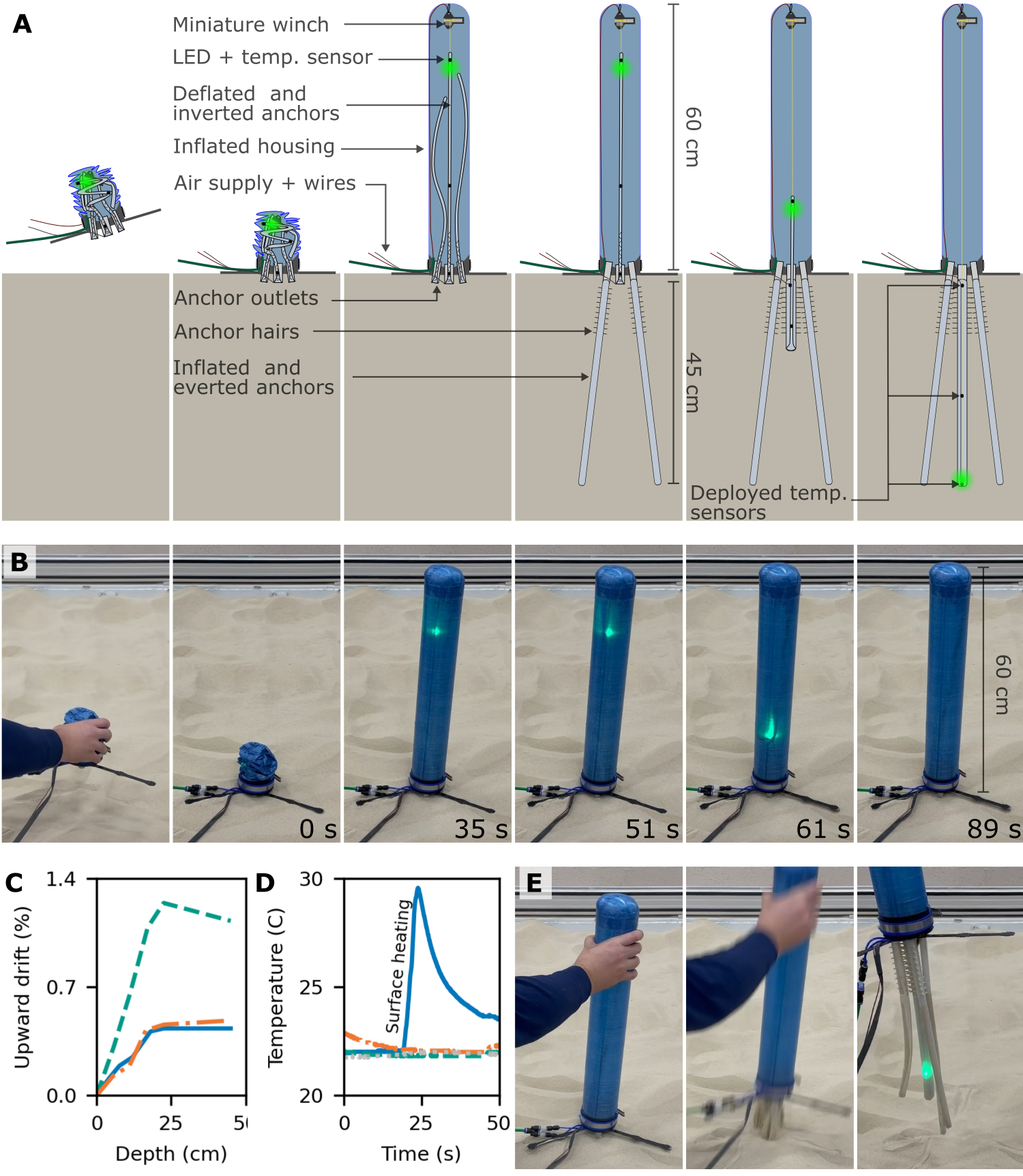}
    \caption{\textbf{Schematics and demonstration of the self-anchoring device} (\textbf{A}) Soft robotic device deployment sequence. First, the compacted device is placed on the regolith; second, the housing is inflated; third, the three primary anchors are deployed; and fourth, the secondary anchor is deployed with the temperature sensors. 
    (\textbf{B}) Demonstration of the device in Martian regolith simulant. (\textbf{C}) The anchor backs out slightly to 1\% or 0.5 cm during the first 22 cm of secondary anchor growth before locking into place, three trials. (\textbf{D}) When a heat gun is applied to the surface of the simulant, the temperature sensor closest to the surface shows an increase in temperature while the deeper ones do not. (\textbf{E}) The 300 g anchor takes approximately 120 N to extract, resulting in a 40:1 anchoring-to-weight ratio.}
    \label{fig:robot_demo}
\end{figure}

Anchoring and sensor deployment are conducted in four steps (Figure \ref{fig:robot_demo}A).
First, the anchoring device is placed on the surface of the ground while attached to an external pneumatic pressure source and electrical interface. 
Second, the device is pressurized to approximately 70 kpa, which inflates the housing and unfurls the anchors inside it.
Third, the pressure is increased to approximately 350 kpa, which extends the self-anchoring primary anchors into the ground. The secondary anchor is prevented from extending by the sequencing winch. 
Fourth, the sequencing winch lets out the string, allowing the secondary anchor to extend into the ground and begin measuring subsurface temperature.
The device is free to move vertically during this anchoring process.

\paragraph{Soft robotic device deployment in martian regolith simulant} 
\label{sec:robot_demo}

We demonstrated the effectiveness of the anchor by deploying it on loose M90 Martian regolith simulant\cite{oravec2021geotechnical} in one of the 2.4$\times$1.5$\times$1 m Traction and Excavation Capabilities Rig (TREC)\cite{creager2025best} bins at NASA Glenn Research Center. 
The device could anchor with only its own weight, needing no externally applied reaction force (Fig. \ref{fig:robot_demo}B). 
As we expected from our terradynamics experiments, we observed the device slightly backing out of the soil during initial deployment of both the primary and secondary anchors before locking into place (Fig. \ref{fig:robot_demo}C shows the effect for the secondary anchor). 
After deploying, we used a heat gun to warm the surface of the simulant around the anchor and observed an increase in temperature at the surface temperature sensor, but not the buried ones (Fig. \ref{fig:robot_demo}D).
The maximum anchoring force for three trials was 108.5 N, 109.5 N, and 139.5 N with an average of 119 N $\pm 17.6$ N.
The weight of the anchoring device is 2.9 N (mass of 300 g), meaning it has an anchoring-to-weight ratio of greater than 40:1.

\section*{DISCUSSION}
In this work, we studied the soft matter interactions in granular media of tip-extending anchors compared to those of rigid intruders. We conducted a series of experiments, including measuring forces and grain velocities during insertion and extraction, exploring how resistive forces were affected as a set of parameters were varied (diameter and angle), and measuring resistive forces in an unconstrained tip extender. We identified critical design principles, and we applied these insights to develop a deployable soft robotic anchor capable of self-anchoring without any external force, generating a 40:1 anchoring-to-weight ratio.

We observed several complex soft robot -- soft matter interactions worthy of further discussion and future study. First, we observed that there is a higher resistance to insertion for the rigid intruders than for the tip extenders (Fig. \ref{fig:int_pull_sand}). Granular intrusion studies indicate that insertion forces are primarily governed by lithostatic pressure acting on the tip and sides of the intruding rod \cite{hamm2011dynamics}. As the intruder penetrates, it must continuously displace and rearrange the particles along its path to create space within the granular medium. In loosely consolidated granular states, this causes particles near the intruder–granular interface to compact. These dynamics were investigated by penetrating a rigid, elongate intruder into granular media, where PIV analysis revealed a reduced rate of particle volume change indicative of material compaction \cite{tapia2013effect}. With increasing depth, the contact area of the intruder enlarges, affecting a greater number of particles within the medium. By contrast, as we demonstrate in (Fig. \ref{fig:PIV}), the interaction between the tip extender-granular media is localized at the tip, while the sides remain in static contact with the surrounding particles. The change in side wall–substrate interaction, from dynamic contact in rigid intrusion to static contact in tip-extension, could account for the greater resistance experienced by the rigid intruder compared to the tip extender. The distinct interaction mechanics between the devices are observable in the evolution of surface deformation during insertion (See supplemental video). 

Second, peak extraction forces were threefold greater for the tip extender than for the rigid intruder, despite their identical geometry. Extraction PIV at the initiation of movement showed reduced mobility of the particles near the tip extender compared to those near the rigid intruder. We propose a hypothesis that may contribute to this reduced particle mobility. The particles surrounding the tip extender may form strong contact networks due to the localized compaction that occurs as the tip advances during insertion. During extraction, these surrounding particles may exhibit substantial resistance to shear, remaining largely stationary with minimal initial movement. In contrast, the particles around a rigid intruder experience global compaction and rearrangement throughout insertion, which may disrupt force networks and reduce their ability to resist extraction.

Third, we observed that our tip extender exhibits zero-reaction-force self-anchoring, but only when the tip extender is free to move vertically. When constrained, the self-anchoring was not observed. This may be due to the motion of the unconstrained extender inducing a jamming transition in the surrounding particles. Its ability to move both upward and downward simultaneously causes opposing forces: while the tip extends deeper, the upper part of the extender moves against gravity. This counter-motion may slightly perturb the surrounding particles, significantly increasing the strength of the granular medium and potentially leading to the development of yield stress \cite{bi2011jamming,behringer2018physics}.
While more work is needed to investigate the origins of this phenomenon, we understand the principle needed for implementing self-anchoring robots: allow the anchor to move in the vertical direction, and the reaction force will drop to zero beyond a critical depth.

Our RFT model predicts some of the above results. Specifically, the model predicts that the net insertion force on a rigid intruder will be higher than on the tip extender. This is because the intruder’s insertion is modeled using both its tip and side walls, whereas the tip extender is modeled using only the tip, which reduces its contact surface area with the substrate. Additionally, the RFT model accurately predicts the extraction force for the unconstrained tip extender that moves in both directions within granular media. This is consistent with the model’s elemental force calculations, which are derived solely from plate motion through granular media and do not rely on prior knowledge of static friction or yield stresses.

We also note that our study has several limitations. Measuring forces on the soft, subterranean body is not trivial.  We measured forces at the base and inferred subterranean forces using modeling and prior results. Future work could involve sensorizing the tip extender, performing  X-ray imaging of particles, or using the Discrete Element Method (DEM) simulation to better understand the distribution of forces between the grains and the soft body. 

Furthermore, our experiments were limited to dry, loose, and relatively homogeneous media, so future work is needed to generalize the findings to other substrates.
We only tested one configuration of hairs, but the design space is very large (e.g., size, shape, material, and spatial density). 
The configuration used showed promise, but future studies could fully explore the trade space.
In PIV experiments, the particles were constrained by two closely spaced walls separated by a distance equal to the device’s diameter. This confinement influences particle motion during device interaction and limits the interpretation of PIV when compared to force data collected in large containers. 
Finally, the anchoring device is not self-contained: the pressure source and electronics are off-board.
Integrating them would increase mass and allow larger anchors to be deployed using only weight as the reaction force.

Our work shows that, from the soft matter physics perspective, a relatively new class of coupled body/substrate interactions can lead to fundamentally distinct granular responses compared to rigid body intrusion. Previous granular intrusion studies have primarily centered on contact with singular rigid bodies, such as penetrometers in civil engineering \cite{sanglerat2012penetrometer} and cannonballs in military applications \cite{robins1805new}. However, the most common interactions in nature with granular media involve complex, multibody shape-changing structures. Both the intruding body and the substrate undergo mutual deformation, leading to reciprocal and emergent stresses. Studies on shape-changing structures/bodies in granular substrate can probe novel and emergent particle reconfiguration and movement in future work.

Overall, this work advances our understanding of the terradynamics of tip-extending anchoring and offers design guidelines for creating anchors that can deploy subsurface sensors with low reaction force and high anchoring force. These lightweight, deployable anchors could have important applications during future missions on the moon, Mars, asteroids, and beyond.

\section*{MATERIALS AND METHODS}

\subsection*{Fabrication of Experimental Devices}

\textbf{Tip extension device}: To fabricate everting tubes, we used Dyneema Composite Fabric. The material was first cut into a rectangular sheet, 15 cm in length, and sized to the desired diameter, with extra length added for overlap. We applied a strip of PSA tape along the edge of the Dyneema and folded it into a cylindrical tube shape with a lap joint. Once the tube was formed, we inverted it and tied an overhand knot to the end to seal it. The remaining end of the tube is sealed around a solid hollow cylinder connected to an air supply. The tail knot’s string extended through a small opening in this cylinder, allowing it to be used for retracting the device. 

\textbf{Rigid intruder}: To build the rod-shaped intruder, we placed a rigid rod with matching dimensions into a Dyneema tube of 1.5 cm diameter. The rod was 3D printed from PLA plastic using a Bambu Lab X1 Carbon 3D printer. 

\textbf{Hairs}: We used urethane casting to create hair-like protrusions. We first 3D printed male molds, then cast them in silicone to create female molds. These were filled with urethane and vibrated to remove air bubbles. Once cured, the urethane parts were carefully adhered to a hairless Dyneema tube. The dimensions of the protrusions are 6$\times$1.5$\times$3 mm$^3$. The vertical distance between individual protrusions was 10 mm. We laid 4 columns of hairs around the periphery of the tube, which were radially separated by 90\degree. 

\subsection*{Insertion and extraction experiments} 
\begin{figure}
    \centering
    \includegraphics[width=.8\linewidth]{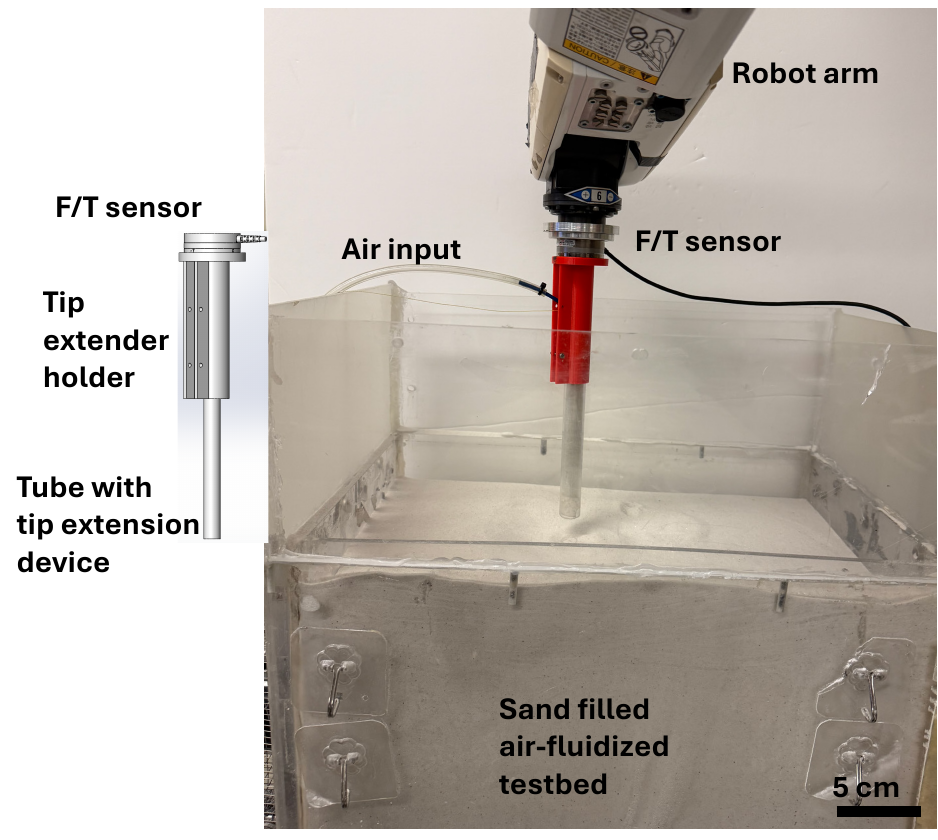}
    \caption{\textbf{Experimental setup used to conduct insertion and extraction experiments of intruders fixed to the force transducer.} All the intruders are inserted 15 cm into fine sand to collect continuous force response recordings. After insertion was complete, they were fully extracted from the sand. }
    \label{fig:SI_robot_arm}
\end{figure}

Sand insertion and extraction experiments were performed in a 42$\times$28$\times$25 cm$^3$ bed (Fig. \ref{fig:SI_robot_arm}) of dry play sand (Sandtastik White Play Sand). Before each experiment, air was introduced through the bottom inlet between trials to fluidize the testbed, reliably returning the sand to a loosely packed initial state with a volume fraction of approximately 58\%. Insertion and extraction experiments were conducted using a DENSO VS087 robotic arm equipped with a 6-axis ATI Mini40 force/torque transducer. Analog force signals were transmitted to a computer via a National Instruments data acquisition system running LabVIEW. We measured the forces in the vertical (z) axes at $1$~kHz during experiments and conducted 5 trials for all penetrating bodies. 

For intruder experiments, the Dyneema tube containing the rigid rod was connected to a hollow extension tube, which was subsequently mounted onto a force transducer via a 20 cm-long holding shaft. Pressurization of the tube was achieved through an air inlet, with airflow regulated using a pressure regulator. During testing, the Dyneema tube containing rod was initially inflated and subsequently inserted vertically to a depth of 15 cm into the sand from the surface using the robotic arm moving at a speed of 1.1 cm/s. Once fully inserted into the sand, we waited 5 seconds to allow the particles to settle, then vertically extracted the rigid intruder at a constant speed of $1~\text{cm/s}$ using the robot arm.

The hairless and hairy tip extension experiments were conducted using the same setup as the rigid intruder experiments. First, the tip of the hollow rigid tube containing the unextended tip extender was positioned at the sand surface. Airflow was then activated and gradually increased as the tip everted and continued to extend through the sand. Once fully extended to a depth of 15 cm, both extenders were vertically extracted at a constant speed of 1.1 cm/s. The displacement is estimated from force-time measurements during pullouts conducted at a constant velocity. The forces are adjusted to start from positive values that correspond to the extraction strength at zero displacement. 

\subsection*{Force-controlled experiments}
\begin{figure}
    \centering
    \includegraphics[width=0.8\linewidth]{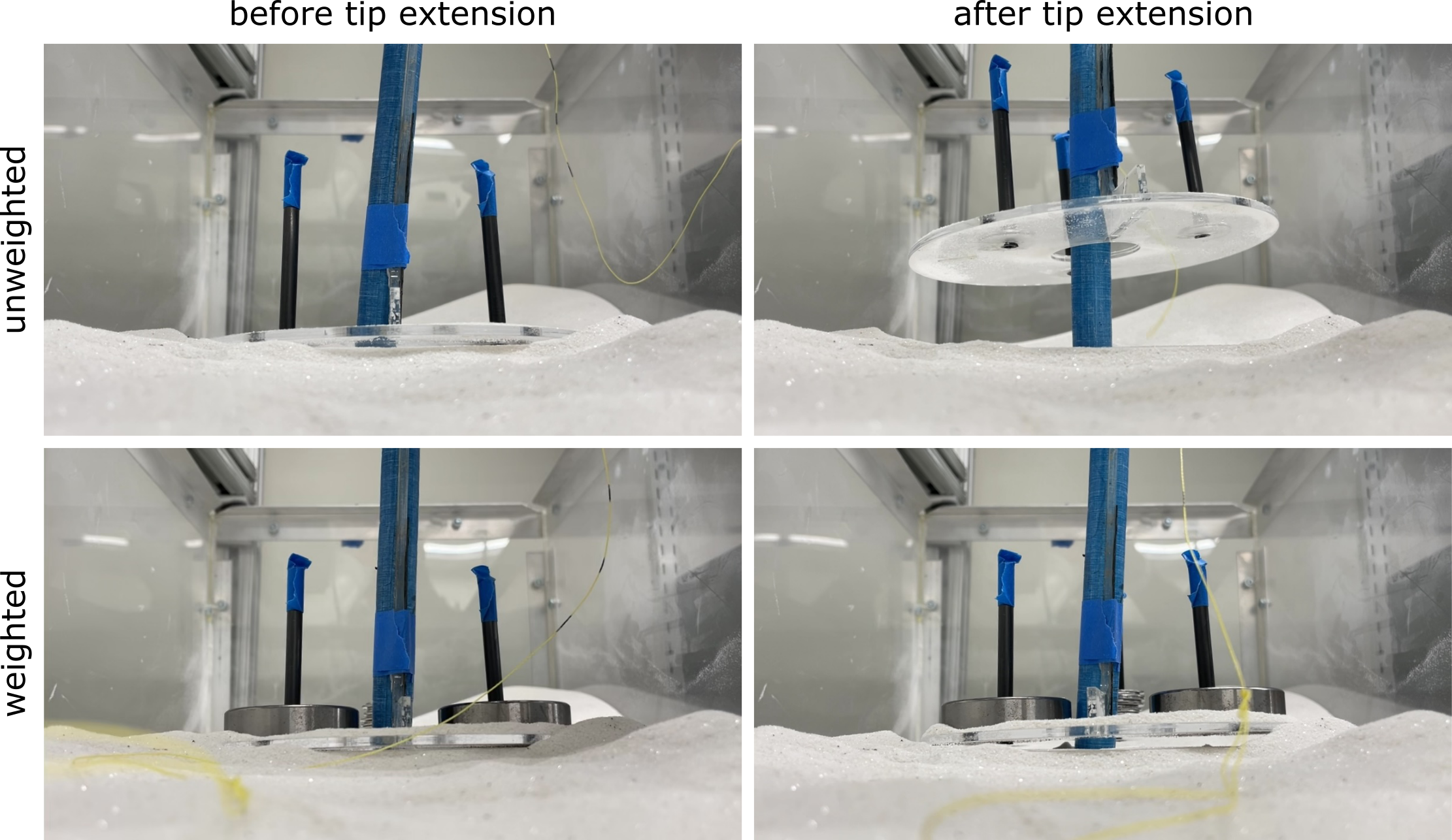}
    \caption{\textbf{Experimental setup used to conduct force-controlled experiments.} The tip-extender was attached to a rigid platform that sat freely on the surface of the sand. Different amounts of weight were placed on it to determine the minimum weight required to prevent the tip extender from backing out of the sand as it grew. The top row shows too little weight, and the tip extender backs out; the bottom row shows just enough weight. The tip extender backs out slightly as it burrows into the sand.}
    \label{fig:SI_weighted_tests}
\end{figure}

The self-anchoring experiments were performed in an air-fluidized testbed measuring 60$\times$30$\times$30 $cm^3$, filled with fine sand as shown in Fig. \ref{fig:SI_weighted_tests}. The tip extension device was mounted onto a load-bearing platform equipped with vertical columns designed to hold weights. We positioned the platform on the sand surface, placed reaction masses onto the columns, and extended the tip extension device to a depth of $3$ cm. Once the device reached this depth, we recorded the maximum mass required to sustain tip extension without failure. We continuously adjusted the mass at each 3 cm depth intervals until the tip extended to a depth where no mass was needed for further growth, i.e., when the body self-anchored. Then the reaction mass values used at each discrete depth were converted to force data. To characterize the extrusion forces of the tip extension device, we first extended it to a prescribed depth. We then extracted the device and measured the resistive force exerted by the surrounding material. Five trials were conducted for each discrete depth.

\subsection*{RFT simulations}
\label{sec:RFT}
\begin{figure}
    \centering
    \includegraphics[width=.8\linewidth]{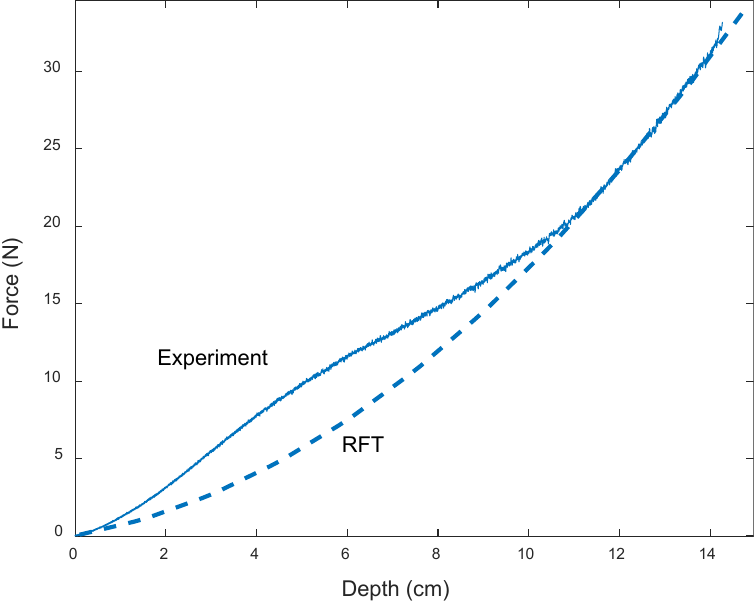}
    \caption{\textbf{Calibration of RFT coefficients to match experiment intrusion into fine sand.}}
    \label{fig:SI_RFT_calibration}
\end{figure}

The self-anchoring simulations based on RFT were performed with MATLAB. We assumed that the insertion force of the tip extender is only governed by the interactions at the tip and modeled it by a thin disc ($1.5$ cm diameter) moving through the granular medium. Similarly, we assumed that the extraction force of the fully extended device is governed only by the side interaction and modeled it as the lateral surface of a cylinder. Consequently, the tip force is $F_t(z)=R_{-zz}(z)\pi r^2$ where $R_{-zz}(z)$ is the resistance on the tip in the downward direction as a function of depth $z$ and $r$ is the radius of the tip extending body. We similarly assumed that side anchoring forces are $F_s(z)=2\pi r\int_0^h R_{xz}(z) dz$ where $R_{xz}(z)$ is resistance on the sides in the upward direction integrated to a depth of $h$. The self-anchoring force is the sum of the insertion and extraction responses of the material. We obtained the resistive forces by integrating the vertical stresses over the moving surface within sand using $F(z)=\int_s{\sigma_z(\beta_s,\gamma_s)|z|_sdA_s}$. Here, $s$ is the leading surface of the moving surface and $dA_s$, $|z|_s$ are the surface area and depth.
$\beta_s$, and $\gamma_s$ are the angle of attack and angle of intrusion of discretized body elements. $\alpha_z(\beta_s, \gamma_s)$ are element stresses per unit depth. To calibrate the RFT coefficients for sand, we used the rigid intruder insertion-depth data as shown in Fig. \ref{fig:SI_RFT_calibration}.

\subsection*{PIV Imaging}
These tests used the identical setup and testbed as the sand insertion and extraction tests presented in Fig. \ref{fig:int_pull_sand}). We positioned intruder and tip extenders against the side wall of the testbed and placed another wall on their opposite side, constraining both the penetrating bodies and particles to create a quasi-2D environment. We used a webcam (Razer Inc.) operating at 30 Hz, positioned to capture the side wall of the testbed. The insertion and extraction experiments were repeated using the penetrating bodies listed in Fig. \ref{fig:int_pull_sand}) against the sidewall. We used MATLAB PIVLab to obtain displacement vectors and velocity fields of the granular interaction images. 

\subsection*{Diameter scaling}
These experiments were conducted in the testbed shown in \ref{fig:SI_vertical_stage}. A vertical actuation stage, consisting of a lead screw and stepper motor, was mounted on the testbed. A force gauge was attached to the vertical stage, which was connected to the hollow tube containing the tip extenders to capture insertion and extraction forces. Tip extenders with diameters ranging from 0.7 cm to 3 cm were all extended to a depth of 15 cm. The tubes containing the extenders were first lowered to the surface of the fluidized sand. Then the tip extension has been initiated by gradually ramping up the air pressure. Once fully extended, we extracted the bodies using the vertical actuation stage. Five trials were conducted for each diameter.

\subsection*{Insertion angle} 
\begin{figure}
    \centering
    \includegraphics[width=.8\linewidth]{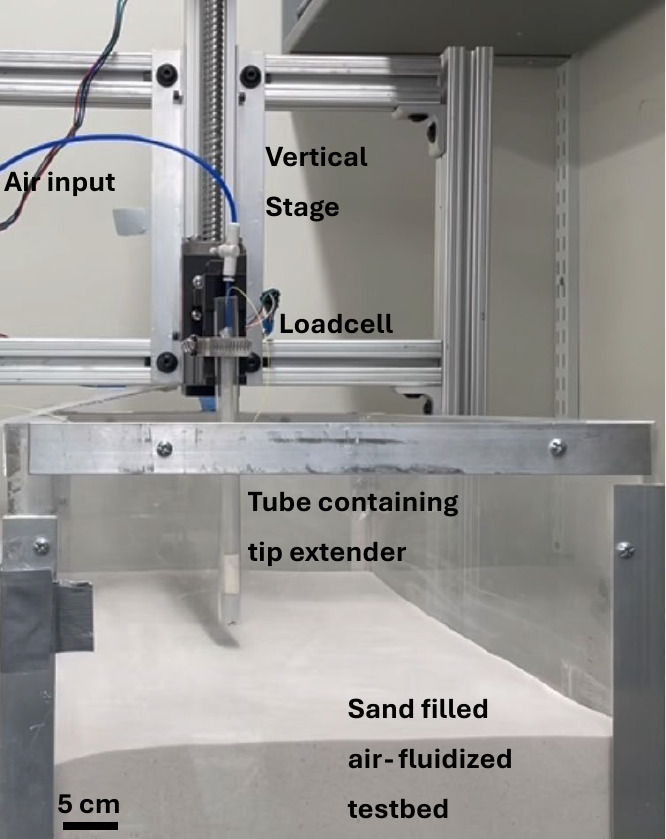}
    \caption{\textbf{Experimental setup used to conduct self-anchoring, diameter variation insertion, and extraction experiments.}}
    \label{fig:SI_vertical_stage}
\end{figure}

\begin{figure}
    \centering
    \includegraphics[width=0.8\linewidth]{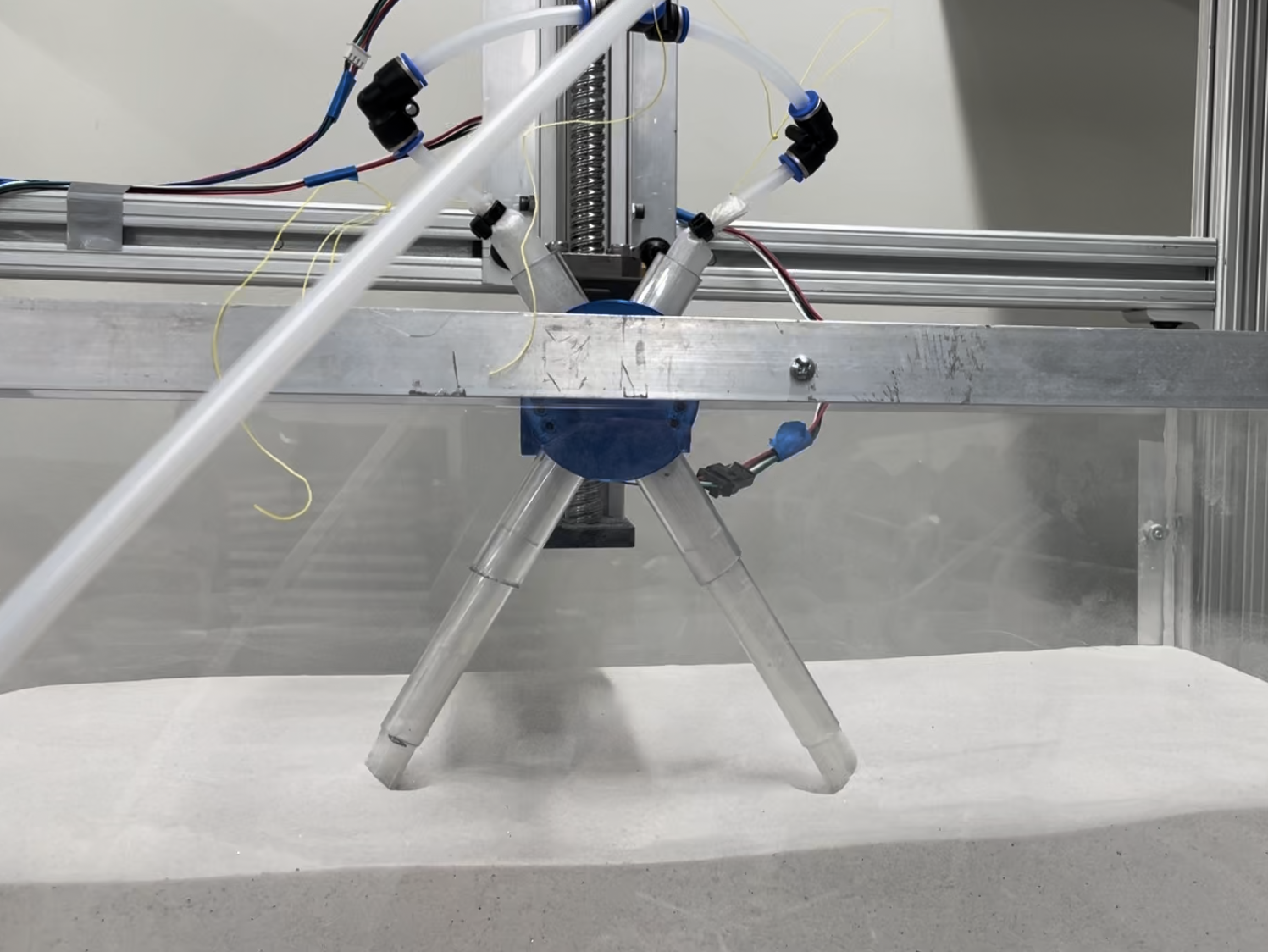}
    \caption{\textbf{Experimental setup used to conduct angled insertion and extraction experiments.}}
    \label{fig:SI_angled}
\end{figure}

These tests used the same testbed as the diameter scaling tests presented in \ref{fig:SI_vertical_stage} but with a different connector with a pair of two identical anchors that could be adjusted for relative angle between them from a parallel and vertical to 60\degree from vertical in an ``X'' shape (\ref{fig:SI_angled}. The insertion angle was adjusted by repositioning the connector in fixed intervals of 15\degree. Both anchors were grown into loose sand for a fixed length of 15 cm, and then pulling both out vertically while measuring force and position. 5 experiments were performed at each angle. 

\subsection*{Self-anchoring soft robotics device fabrication and testing}
The inflatable body and anchors of the device were made of a laminated film comprising thermoplastic polyurethane (TPU) and Dyneema composite fabric (ultra high molecular weight polyethylene (UHMWPE) fibers sandwiched between polyester film). A 50 µm thick TPU (American Polyfilm MT 2001) bladder of the desired diameter was heat-welded together and then covered in a 50 µm thick Dyneema composite fabric (0.51 oz/yd$^2$) with 50 µm thick pressure sensitive adhesive (M3 VHB F9460PC) with a 130 mm wide lap joint. The inflatable anchor housing was made the same way but with 100 µm thick Dyneema composite fabric (1.0 oz/yd$^2$) and a 5 cm wide lap joint. Hairs were cast in urethane rubber (VytaFlex 60, Smooth-on) directly onto Dyneema composite fabric in a silicone mold made from a 3D printed parent mold. 
The rigid base was 3D printed in nylon (Onyx Black, MarkForged). The temperature sensors were 1-wire digital thermometers (UMW DS18B20).

\paragraph{Self-anchoring device experiments}
The anchoring experiment was performed in loose M90 Martian regolith simulant. Before each trial, the 1 m$^2$ region around the test was mixed with an auger for a loose compaction state. The maximum anchoring force was measured by attaching a digital force gauge (Mark-10 M3) to the top of the inflatable housing and manually pulling until the device broke free of the ground.

\bibliography{scirefs} 
\bibliographystyle{sciencemag}

\section*{Acknowledgments}
Funding: This work was partly supported by NASA Early Career Faculty Award 80NSSC21K0073, NASA Presidential Early Career Award in Science and Engineering (PECASE) 80NSSC25K0289, and National Science Foundation grant 1944816. 
Daniel I. Goldman and Deniz Kerimoglu acknowledge support from NSF PoLS, NSF FRR and JPL.
The work of N. Naclerio was funded in part by a NASA Space Technology Research Fellowship.
N. Naclerio, E. Hawkes, and D. Goldman are inventors of two patent applications related to this project.

\section*{Supplementary Materials}

\section*{Supplementary Text}

\subsection*{The effects of media on insertion and extraction}
We examined how insertion and extraction forces were influenced by the variations in particle friction, size, and cohesion. Specifically, we replaced high-friction, small-grain sand with low-friction, large-grain glass beads, as well as with powder media to incorporate cohesion between particles. Experiments were conducted in cylindrical testbeds—17 cm in diameter and 25 cm in height for glass bead trials using 3 mm spherical particles at a volume fraction of 0.5, and 20 cm in diameter and 25 cm in height for powder trials using dry cornstarch (particle size: 10–20 µm) as a model cohesive substrate. We tapped and shook the powder testbed to achieve a closely packed state, resulting in a measured bulk volume fraction of $\phi = 0.56$. This value was obtained by calculating the ratio of solid volume to occupied volume, using the density of cornstarch particles ($1.34~\text{g/mL}$) \cite{fuentes2019fractionation}. We used the identical penetrating bodies and followed the insertion and extraction procedures in sand. Five trials were conducted for the glass beads and powder experiments.

\begin{figure} [!h]
    \centering
    \includegraphics[width=.9\linewidth]{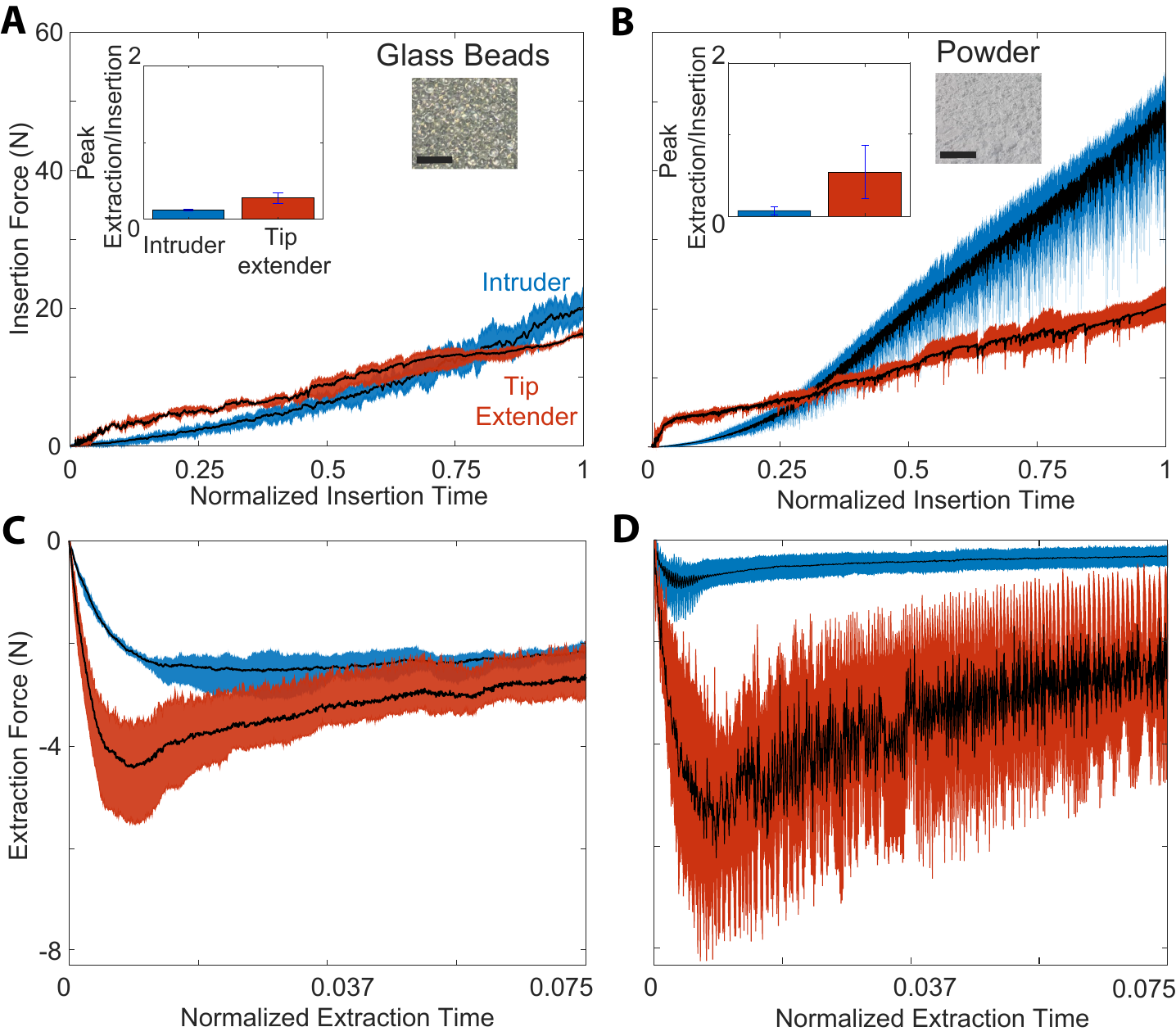}
    \caption{\textbf{The effect of media on intrusion and extrusion forces.} The scale bar represents 1 cm. The error bars represent the lower and upper bounds of the peak extraction-to-insertion force ratios. (\textbf{A}) The Insertion force in glass beads is similar for both devices. The average insertion times were 13 s for the rigid intruder and 17.3 s for the tip extender. The mean standard deviations of the shaded regions for the intruder and tip extender are 0.90 N and 0.73 N, respectively. (\textbf{B}) In powder, the insertion force for the intruder is larger. The average insertion times were 9 s for the rigid intruder and 10 s for the tip extender. The mean standard deviations of the shaded regions for the intruder and tip extender are 2.34 N and 1.12 N, respectively.
    (\textbf{C}) Extraction force in glass beads is larger for the tip extender than for the intruder. (\textbf{D}) In powder, this difference is much larger. The average extraction time for all the devices was 13.5 s. The mean standard deviations of the shaded regions are 0.20 N for the intruder and 0.61 N for the tip extender in glass beads, and 0.17 N and 1.43 N, respectively, in powder.}
    \label{fig:2_media}
\end{figure}

The insertion forces on the rigid intruder and tip extender within glass beads (low-friction and large-size particles) were similar (Fig. \ref{fig:2_media}A), unlike in sand (Fig. \ref{fig:int_pull_sand}A). In contrast, the insertion force difference between devices in powder (high-friction and small-sized cohesive particles - (Fig. \ref{fig:2_media}B)) is larger. This difference in insertion forces could be related to interparticle frictional effects between the two media. Additionally, the tip extender’s insertion forces in glass beads and powder, both closely packed, were significantly higher than those in sand, which was loosely packed. This can be attributed to the limited space for the particles to rearrange near the tip, leading to increased resistance. 

For the extraction experiments, we observe that the forces are consistently larger for the tip-extending device than for the rigid intruder in all the media tested (Fig. \ref{fig:2_media}C-D and Fig. \ref{fig:int_pull_sand}B). However, the extraction-to-insertion ratio for the glass beads and powder was significantly lower than in sand. Similarly, this could be attributed to the tip extension compacting the loosely packed sand during insertion, whereas the closely packed beads and powder resist compaction. Moreover, as particle–particle friction increases, the difference in extraction forces between the intruder and the tip extender becomes more pronounced. This indicates that higher interparticle friction amplifies the material resistance during extraction, particularly for the tip extension case.

Insertion of the intruder into the powder exhibited noticeable oscillatory forces compared to the tip extension case. These oscillations can be attributed to the stick–slip phenomenon commonly observed during intrusion into closely packed granular media \cite{hamm2011dynamics}. In contrast, tip extension appeared to alter this behavior, resulting in smoother penetration. During extraction, however, the trend reversed—the tip extender exhibited more pronounced oscillatory forces than the intruder.

\subsection*{PIV analysis of the hairy tip extender}
\begin{figure}
    \centering
    \includegraphics[width=.8\linewidth]{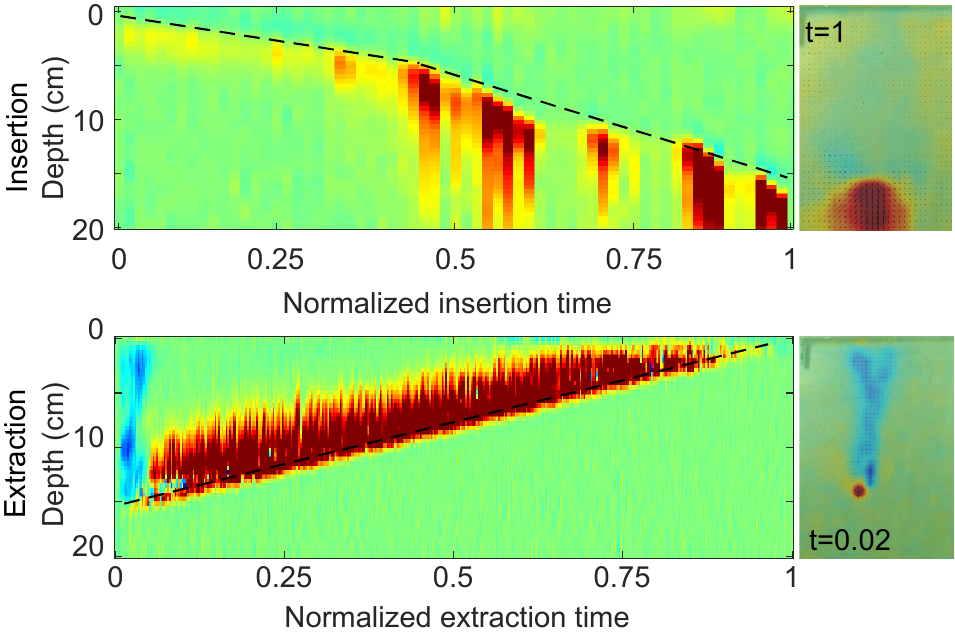}
    \caption{\textbf{Heatmap of hairy tip extender showing particle velocities}. PIV images of the hairy tip extender during insertion show similar trends to those of the hairless tip extender. Extraction PIV reveals a slightly delayed onset of particle motion, similar to the hairless tip extender, but exhibits a more pronounced downward material flow.}
    \label{fig:SI_PIV_hair}
\end{figure}

We performed PIV analysis on the insertion and extraction motions of the hairy tip extender (Fig. \ref{fig:SI_PIV_hair}). We observed that adding hairs further increased particle movement during both insertion and extraction. During insertion, the increased particle mobilization induced by the hairs may have been offset by improved anchoring along the hairy sidewalls, potentially accounting for the comparable bulk insertion forces between hairless and hairy extender (Fig. \ref{fig:int_pull_sand}). During extraction, the hairy extender engages a larger region of particles compared to the other devices, as observed by the bluer region at the onset of extraction. This is consistent with the increased peak extraction forces for the hairy tip extender.

\section*{Supplementary Figures}

\begin{figure}
    \centering
    \includegraphics[width=.8\linewidth]{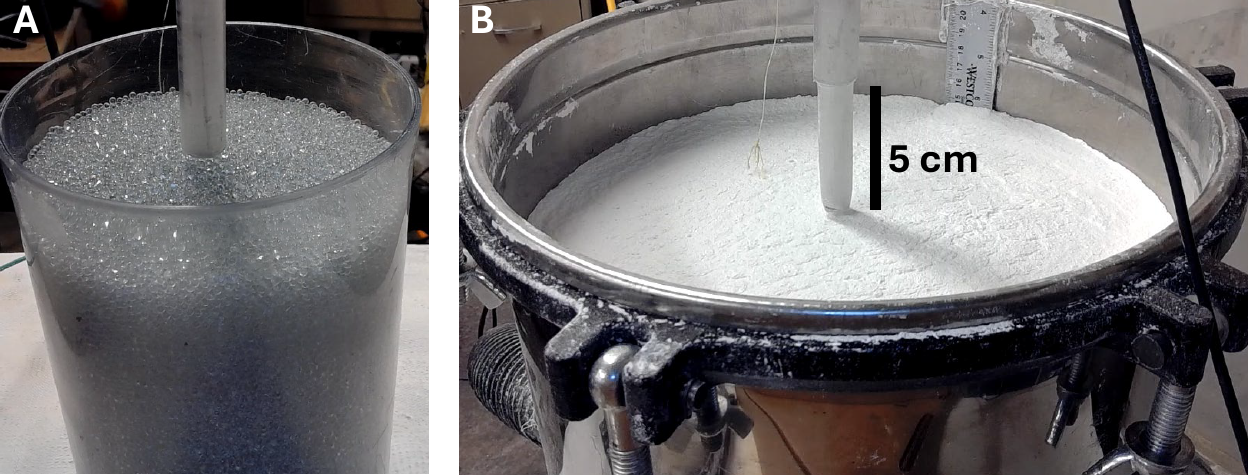}
    \caption{\textbf{Apparatuses used for A) Glass bead experiments and B) Powder experiments.}}
    \label{fig:SI_powder_setup}
\end{figure}

\clearpage

\end{document}